\documentclass[10pt,twocolumn,letterpaper]{article}

\usepackage{iccv}
\usepackage{times}
\usepackage{epsfig}
\usepackage{graphicx}
\usepackage{amsmath}
\usepackage{amssymb}

\usepackage{cuted}
\usepackage{capt-of}
\usepackage{booktabs}
\usepackage{authblk}

\usepackage[breaklinks=true,bookmarks=false]{hyperref}

\iccvfinalcopy 

\def\iccvPaperID{****} 

\ificcvfinal\fi

\begin{document}

\title{Towards Adversarially Robust and Domain Generalizable Stereo Matching \\by Rethinking DNN Feature Backbones}

\author{Kelvin Cheng}
\author{Christopher Healey}
\author{Tianfu Wu}
\affil{North Carolina State University}

\makeatletter
\let\@oldmaketitle\@maketitle
\renewcommand{\@maketitle}{\@oldmaketitle
  \includegraphics[width=0.98\textwidth]{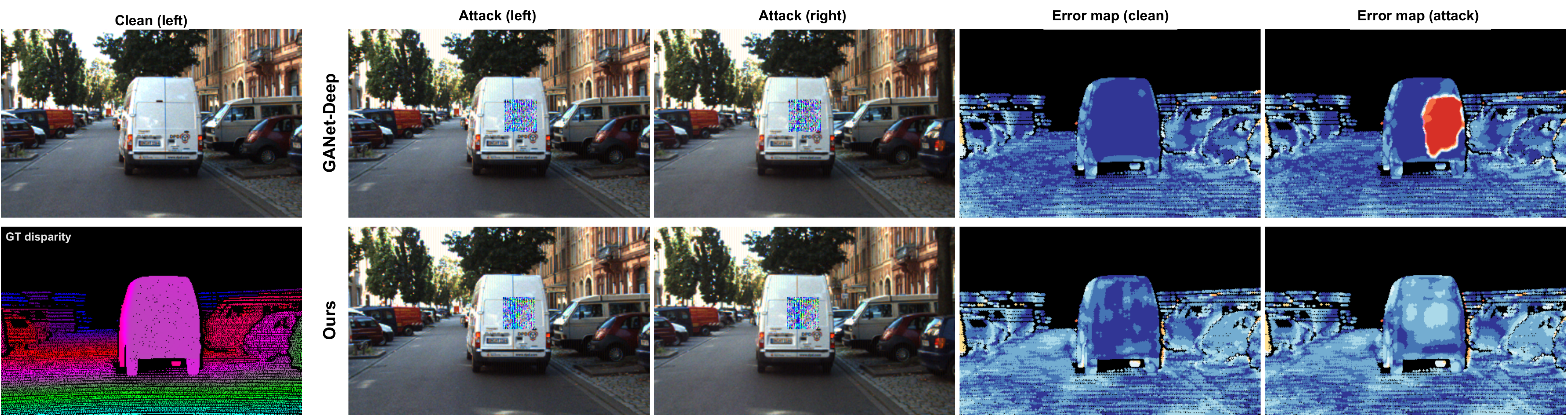}\captionof{figure}{Examples of attacking stereo matching in the KITTI2015~\cite{Menze2015CVPR} dataset. GANet-Deep~\cite{GANet} results on the top row, our results on the bottom row. The attack is based on the proposed stereo-constrained projected gradient descent (PGD) attack within a patch, which by design preserves the photometric consistency of non-occluded regions. One of the state-of-the-art methods, GANet-Deep shows a significant drop in performance (the last column), while the proposed method shows much stronger resistance to the attack. \label{fig:teaser}}\bigskip}
\makeatother

\maketitle
\ificcvfinal\thispagestyle{empty}\fi

\begin{abstract}
Stereo matching has recently witnessed remarkable progress using Deep Neural Networks (DNNs). But, how robust are they? Although it has been well-known that DNNs often suffer from adversarial vulnerability with a catastrophic drop in performance, the situation is even worse in stereo matching. 
This paper first shows that a type of weak white-box attacks can overwhelm state-of-the-art methods. The attack is learned by a proposed stereo-constrained projected gradient descent (PGD) method in stereo matching.
This observation raises serious concerns for the deployment of DNN-based stereo matching. 
Parallel to the adversarial vulnerability, DNN-based stereo matching is typically trained under the so-called simulation to reality pipeline, and thus domain generalizability is an important problem.
This paper proposes to rethink the learnable DNN-based feature backbone towards adversarially-robust and domain generalizable stereo matching by completely removing it for matching. In experiments, the proposed method is tested in the SceneFlow dataset and the KITTI2015 benchmark, with promising results. We compute the matching cost volume using the classic multi-scale census transform (i.e., local binary pattern) of the raw input stereo images, followed by a stacked Hourglass head sub-network solving the matching problem. It significantly improves the adversarial robustness, while retaining accuracy performance comparable to state-of-the-art methods. It also shows better generalizability from simulation (SceneFlow) to real (KITTI) datasets when no fine-tuning is used. 
\end{abstract}

\section{Introduction}

\begin{figure*} [t]
\begin{center}
\includegraphics[width=0.99\linewidth]{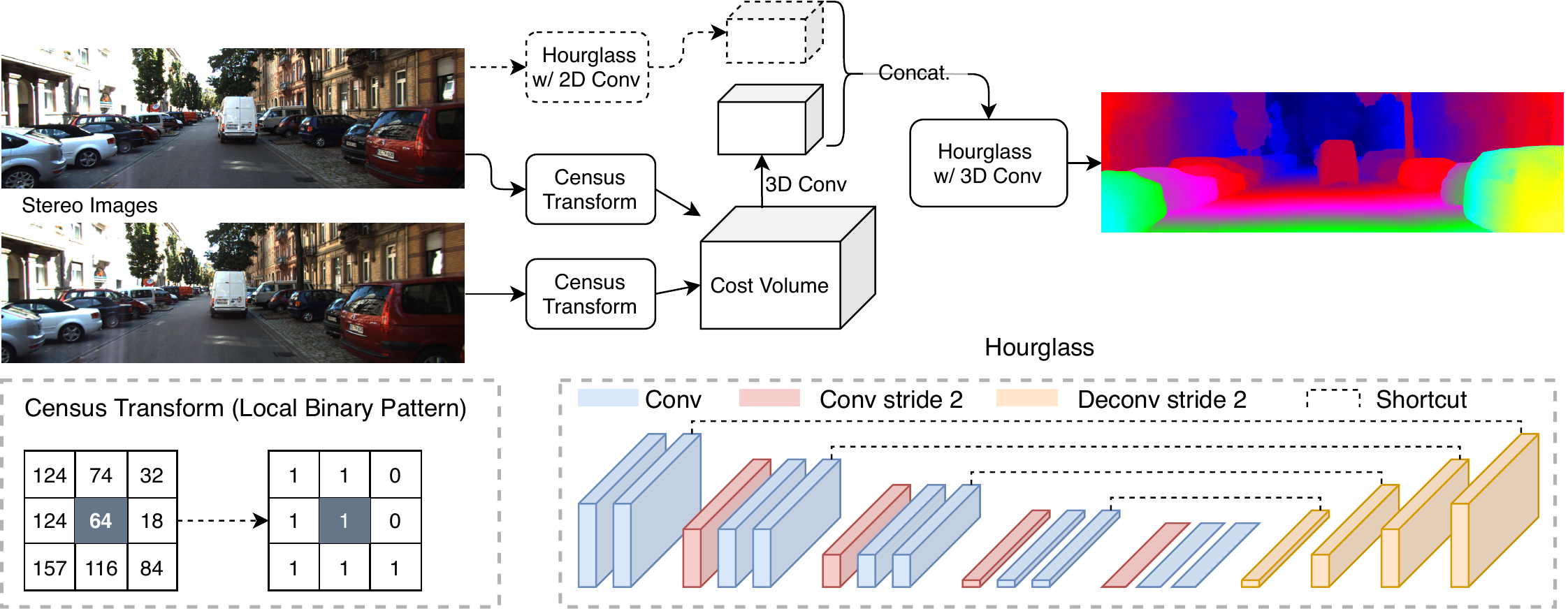}
\end{center}
   \caption{Illustration of the proposed minimally-simple workflow for stereo matching. The key difference between the proposed method and prior art lies in the way of computing the cost volume. The proposed method harnesses the classic multi-scale census transform (left-bottom) of raw intensity of an input stereo image pair, while prior art utilize features computed by a ConvNet feature backbone on an input stereo image pair. The proposed method also exploits ConvNet features computed only using the left reference image, as contextual information to the cost volume. {Note that we also test the workflow without using the ConvNet feature context branch, that is to completely remove the ConvNet feature backbone.} For the cost aggregation component, the proposed method utilizes a stacked Hourglass sub-network equipped with 3D convolution. Please see the text for detail.}
\label{fig:model}
\end{figure*}

Stereo matching remains a long-standing problem in computer vision that has been studied for several decades. It has great potential in a wide range of applications such as autonomous driving and robot autonomy.  

As in many other computer vision problems, Deep neural networks (DNNs) have made tremendous progress in stereo matching. 
The growing ubiquity of DNNs in computer vision dramatically increases their capabilities, but also increases the potential for new vulnerabilities to attacks~\cite{Segmentation&DetectionAttack,GenerativeModelAttack,FaceAttack,HotFlip}. This situation has become critical as many powerful approaches have been developed where imperceptible perturbations to DNN inputs could deceive a well-trained DNN, significantly altering its prediction. Such results have initiated a rapidly proliferating field of research characterized by ever more complex attacks~\cite{FGSM, PGD, PracticalBlack-BoxAttacks,EnsembleAttack,TransferabilityAttack1, TransferabilityAttack2,AdvDistillation} that prove increasingly strong against defensive countermeasures~\cite{LogitPairing,FeatureDenoising,distillation_defense}. {For the trade-off between accuracy and adversarial vulnerability, DNNs seem to have become the Gordian Knot in state-of-the-art computer vision systems.} 

Since stereo matching methods are widely used in autonomous driving, adversarial vulnerabilities in these models can lead to catastrophic consequences. \cite{wong2021stereopagnosia} test attacks on stereo images independently, resulting in perturbations that will alter the colors of the corresponding projections of the same physical point and thus may not be realizable and threatening in practice. To find out whether stereo matching methods are vulnerable in a physically realizable setting, we propose the stereo-constrained projected gradient descent (PGD) attack and show that state-of-the-art methods are indeed vulnerable even when the color differences between corresponding pixels are preserved.

To defend against adversarial attacks, most methods rely on adversarial training~\cite{PGD}, which may suffer from decreasing performance, long training time, and over-fitting to specific attacks and datasets. In contrast, we propose to utilize domain-specific knowledge to facilitate the built-in robustness of the neural networks. Because of the strong photometric consistency between stereo images, stereo matching provides an ideal case to defend against adversarial attacks through the design of the neural network. For non-occluded regions in stereo images, the corresponding pixels of the same physical point have similar colors. We suspect that by using DNN features for matching, attacks will increase the matching costs for features that belong to the same physical point. Therefore, we propose to remove DNN features for matching and use hand-crafted features that will preserve the low color differences for true pairs. To make the cost as hard to alter as possible, we use local binary patterns that compare each pixel intensity to their neighbors (i.e., Census Transform~\cite{hirschmuller2008evaluation,bleyer2010does}) as the feature descriptor. On the other hand, since DNN features are useful for high-level semantic information that will facilitate the estimation of occluded and textureless regions, we use a feature backbone for the reference image only to contextualize the input. The non-parametric cost volume and the contextual information will be fed through a head sub-network playing the role of a learnable optimizer that seeks the best matching result (Fig.~\ref{fig:model}). In essence, we cast stereo matching as a cost aggregation/optimization problem over a non-parametric cost volume. In experiments, we show that this more transparent approach improves adversarial robustness significantly while maintaining high accuracy.

Parallel to the adversarial vulnerability, \textit{cross-domain generalizabilty} also is an important problem in stereo matching: DNN-based stereo matching is typically pre-trained under the so-called simulation to real (Sim2Real) pipeline due to the high cost of collecting ground-truth matching results in practice and the data-hungry aspect of DNNs. It has been shown that DNNs may learn shortcut solutions that are strongly biased by the training dataset~\cite{geirhos2020shortcut}. Removing the DNN feature backbone for matching will induce the DNN to be a more general cost volume optimizer, thus alleviating the opportunity of shortcut learning in the feature space and resulting in better performance in cross-domain deployment, especially when no fine-tuning is used. These are verified in our experiments from the SceneFlow dataset~\cite{dispnet16} to the KITTI benchmark~\cite{Menze2015CVPR} when no fine-tuning is used.

\section{Related Work and Our Contributions}
\textbf{Deep Stereo Matching.}
After \cite{StereoPatch} developed the first deep learning approach for stereo matching, \cite{dispnet16} built the first end-to-end trainable DNN-based method DispNet and constructed SceneFlow, a large-scale synthetic dataset containing around $40,000$ images. In GCNet, \cite{gcnet2017} further extend the end-to-end approach by concatenating features in the cost volume stage, using 3D convolutional layers for cost aggregation, and introducing the soft $\arg\min$ operator to compute the expected disparity. Most subsequent approaches followed these design choices and use SceneFlow for pretraining~\cite{Menze2015CVPR}.

 \cite{chang2018pyramid}, \cite{AMNet}, and ~\cite{GroupWiseStereo} further improve the cost aggregation stage. Chang \etal proposed to use a Spatial Pyramid Pooling (SPP) Module for feature extraction and to use the stacked Hourglass structures~\cite{hourglass_2016} for the cost aggregation. 
 \cite{Yee_2020_WACV} speed up stereo matching by using highly optimized hand-crafted features (e.g. Census Transform and Sum of Absolute Differences). Hourglass and Census Transform are also used in our approach.

 \cite{GANet,SPNetStereo} proposed to propagate cost spatially to reduce the number of 3D convolutional layers. 
 \cite{LEAStereo} applies Neural Architecture Search (NAS) techniques to automatically find optimal architectures for each stage and further improve the performance. These approaches are the current state-of-the-art in the KITTI 2015 benchmark~\cite{Menze2015CVPR}.

\textbf{Adversarial Attacks and Defense.}
Assuming full access to DNNs pretrained with clean images, white-box targeted attacks are powerful ways of investigating the brittleness of DNNs. Many white-box attack methods focus on norm-ball constrained objective functions~\cite{LBFGS,IFGSM,CWAttack,MIFGSM,DDNAttack}. 
By introducing momentum in the MIFGSM method~\cite{MIFGSM} and the $\ell_{p}$ gradient projection in the PGD method~\cite{PGD}, they usually achieve better performance in generating adversarial examples.

In autonomous driving, physically realizable attacks are investigated in many tasks~\cite{Eykholt_2018_CVPR, ranjan2019attacking, Cao_2020_MLSys, tu2020physically, tu2021exploring}, except for stereo matching. Although \cite{wong2021stereopagnosia} show that DNN-based stereo matching methods are vulnerable against unconstrained adversarial attacks on both images separately, without enforcing photometric consistency, these attacks will violate the underlying physical properties of binocular vision and thus are not realizable in practice. As a result, unconstrained attacks cannot compute adversarial patches to fool stereo systems. Therefore we intentionally design the stereo-constrained PGD attack to further investigate the adversarial robustness in more realistic settings with the presence of photometric consistency. 
\cite{ranjan2019attacking} studied adversarial attacks in optical flow, which is inherently easier to attack than stereo matching due to the problem difficulty. By leveraging the insights from Ranjan's work, our work may shed light on studying more robust optical flow networks.

Towards defense, adversarial training is the most widely used method to improve adversarial robustness~\cite{PGD,Bai_ijcai2021-591}. However, it also suffers from the disadvantages of dropping accuracy, long training time, and over-fitting to specific attacks and datasets. While adversarial training is universal to all kinds of DNNs, our method increases the built-in robustness by utilizing the photometric consistency of stereo matching, thus avoiding the mentioned disadvantages. It can also be combined with adversarial training to further improve robustness.

\textbf{Our Contributions.} This paper makes three main contributions to the field of stereo matching:  
\begin{itemize}
    \item It proposes a novel design for stereo matching, which shows significantly better adversarial robustness and cross-domain (Sim2Real) generalizability when no-fine tuning is used.
    \item  It presents the stereo-constrained projected gradient descent (PGD) attack method, which by design preserves photometric consistency to show the more serious vulnerabilities of state-of-the-art DNN-based stereo matching methods.
    \item  It showcases a deep integrative learning paradigm by rethinking the end-to-end DNN feature backbones in stereo matching, which sheds light on potentially mitigating shortcut learning in DNNs via leveraging classic hand-crafted features if a problem-specific sweet spot can be identified (such as the cost volume in stereo matching). 
\end{itemize}

\section{Approach}
In this section, we present the proposed method and the stereo-constrained PGD attack method to evaluate the brittleness of DNN-based stereo matching methods.

\subsection{The Proposed Method}
As illustrated in Fig.~\ref{fig:model}, the proposed workflow consists of three main components as follows.

\textbf{i) Computing the Cost Volume Using Multi-Scale Census Transform.}
Most current stereo matching methods use DNN-based features to form the 4D cost volume. In terms of matching, DNNs can increase the uniqueness of the feature for each pixel, but they also suffer from the inherent adversarial vulnerability. In contrast, traditional methods often use simple window-based similarity functions to initialize the costs, then rely on the optimization or cost aggregation stage to integrate all local cost information~\cite{szeliski2010computer}. Following the same philosophy, we propose to use hand-crafted feature descriptors and similarity functions that are less sensitive to adversarial perturbations to initialize the costs, then rely on DNNs to integrate the local cost information. Specifically, we want the feature descriptor to change as little as possible when local intensities are perturbed. This specific requirement lead us to the Census Transform, a traditional feature descriptor that is developed to eliminate the issue of radiometric differences caused by different exposure timing or non-Lambertian surfaces. Previous studies find that Census Transform is the most robust and well-rounded cost function with global or semi-global methods~\cite{hirschmuller2008evaluation,bleyer2010does}.

We use grey-scale raw intensity values in computing the census transform. Given a local window patch $W$ centered at a pixel $u\in\Lambda$, the census transform computes the local binary pattern (the left-bottom in Fig.~\ref{fig:model}) by comparing each neighboring pixel $v\in{W}$ with $u$ such that it equals $1$ if $I(v)>=I(u)$ and $0$ otherwise. Hamming Distance (i.e. the number of different values in two bit strings) is used to compute the cost between two patches.

Unlike in traditional semi-global or global methods in which the cost of each pair can only be a scalar, we take advantage of the flexibility of DNNs and design the multi-scale census transform to incorporate the context at different scales. Specifically, We use local windows with sizes from $k_1$ to $k_2$ (e.g. $k_1=3, k_2=11$ in our experiments) so there are $K=k_2-k_1+1$ (e.g., 9) costs associated with each matching candidate pairs. To normalize the cost at each scale, we divide the Hamming Distance by the number of pixels of each local window. 
For an input stereo image pair, $I^L$ and $I^R$ with the spatial dimensions $H\times{W}$, assume the maximum disparity level denoted by $\ell$, the initial cost volume is a 4-D tensor of the size $H\times{W}\times{\ell}\times{K}$. To reduce the computational cost, we use 3D convolutions to down-scale the cost volume to be $1/3H\times{1/3W}\times{1/3\ell}\times{C}$, where $C=32$ is the number of channels, as typically done by prior art.

\textbf{ii) Contextualizing the Cost Volume and Aggregating the Cost.}
Although being robust to adversarial attacks, the census transform based cost volume alone is not sufficiently powerful to handle occlusion and more challenging semantic information, such as transparent objects and specular reflections. 
We introduce a 2-stack Hourglass module with 2D convolutions to extract context information from the left reference image, resulting in a $1/3H\times{1/3W}\times{C}$ feature map which is unsqueezed along the second dimension (i.e., copying the feature map $1/3\ell$ times ) to form a same size tensor as the down-scaled cost volume. The two are then concatenated along the second last dimension.

The contextualized cost volume will be fed into a 3-stack Hourglass module with 3D convolutions for the cost aggregation stage.

\textbf{iii) Disparity Map Prediction and the Loss Function.}
To predict the final disparity map $D(u),\forall u\in \Lambda$, the output of each stack in the Hourglass module of the cost aggregation is first up-sampled to the original size $H\times{W}\times{\ell}$, denoted as $D_s(x,y,d)$ where $s$ is the stack index in the stacked Hourglass module. Then, similar to the method used in~\cite{gcnet2017}, the predicted disparity map $D_s(x,y)$ is computed by,
\begin{equation}
    D_s(x,y)= \sum_{d=1}^{\ell}{d\times{Softmax(D_s(x,y,d))}}, 
\end{equation}
where $Softmax$ is applied along the last dimension in $D_s(x,y,d)$. 

In training, we use the smooth $L_1$ loss due to its robustness at disparity discontinuities and low sensitivity to outliers~\cite{chang2018pyramid,GANet}. Given the ground-truth disparity map $D^*(u)$, the loss is defined by, 
\begin{equation}
    Loss(\Theta; D^*) = \sum_{s=1}^S \beta_s \cdot \frac{1}{|\Lambda|} \sum_{u\in \Lambda} Smooth_{L_1}(D_s(u)-D^*(u)),
\end{equation}
where $\Theta$ collects all parameters in our model, $\beta_s$ represents the weight for the output from a stack $s$ (e.g., 0.5, 0.7, and 1 are used for the 3-stack Hourglass module in our experiments), $u=(x,y)\in \Lambda$, and the smooth $L_1$ function is defined by,
\begin{equation}
Smooth_{L_1}(z)=
\begin{cases}
\frac{z^2}{2}, & \text{if}\ z<1 \\
|z|-0.5, & \text{otherwise.}
\end{cases}
\end{equation}

\section{Stereo Constrained PGD Attacks} \label{sec:attacks}
To study the brittleness of DNN based stereo matching models, we intentionally develop a realizable attacking method based on the PGD method~\cite{PGD}, which retains the underlying photometric consistency in stereo matching by changing the intensities of the same physical point in both images. More specifically, in learning attacks, the same amounts of perturbations are added to each pair of correspondence pixels in the left and right images simultaneously while occluded areas will not be modified. Since the left image is the reference image for computing the disparity loss, we disallow to attack and evaluate occluded regions of the reference image, which prevents the perturbation to attack the regions where the estimation does not rely on matching.

Given a perturbation map $P(x,y), \forall (x,y)\in \Lambda$, the distorted intensities for each pixel location $(x,y)$ are computed as:
\begin{equation}
    \begin{split} 
    &I_{adv}^L(x,y) = I^L(x,y) + P(x-D(x,y),y), \\
    &I_{adv}^R(x,y) = I^R(x,y) + P(x,y), \label{eq:attack}
    \end{split}
\end{equation}
where $D(x,y)$ is the ground-truth disparity map. 

Consider two corresponding patches on the left and the right images containing the same physical points, the absolute sum of difference between these two patches will remain the same after the attack.

We use the $L_{\infty}$ norm to measure similarities between images. Two images will appear visually identical under a certain threshold. To learn a $L_{\infty}$ bounded adversarial perturbation $P^{adv}$, the iterative PGD method is used,
\begin{equation}
    P_{t+1}^{adv} = clip_{P}^{\epsilon}\{P_{t}^{adv}+\alpha\cdot {sign(\nabla_{P}L(P_{t}^{adv}))}\},
\end{equation}
where $t=1, 2, \cdots, T$ and $P^{adv}_0$ starts with all zeros. $L$ denotes the mean absolute error between the predicted disparity map for the perturbed images and the ground-truth disparity map. And, $clip_{P}^{\epsilon}$ clips the perturbation to be within the $\epsilon$-ball of the corresponding zero-plane and the maximum color range. Throughout our experiments, we set $\epsilon=0.06$ or $0.03$, $\alpha=0.01$, and $T=20$.

\textbf{Attack Census Transform.}
Since Census Transform contains the non-differentiable comparison operator, the gradients from the constructed cost volume cannot be back-propagated directly to the input images thus leading to an illusion of safety, \ie the obfuscated gradient problem \cite{obfuscated_gradients}. For fair comparisons with differentiable methods, we combine subtraction and the sigmoid function as a differentiable approximation of the comparison operator. 
\begin{equation}
    a>b \approx sigmoid(a-b)\cdot{C}
\end{equation}
We use a large constant (\ie $C=10^5$) such that the output of the sigmoid function is close to either zero or one. Without using this differentiable approximation, our method without the contextual feature backbone will be \textbf{unattackable} since the gradient flows are completely blocked.   

\textbf{Robustness of Census Transform.}
From the perspective of attacks, the binary patterns generated by Census Transform is more difficult to alter due to the comparison operator.  Given a threshold of maximum pixel difference in perturbation, \textbf{neighbors will not be altered if their difference with the center is larger than twice the threshold.} If the attack does not violate photometric consistency, it will be even harder to alter the cost between binary patches of corresponding pairs. Specifically, if a neighboring pixel appears in both the left and the right binary patches, its relative magnitude relationship with the center pixel will be the same for both patches, no matter how its intensities change. It is our interest to test if this highly non-linear operator can defend the DNNs against attacks.

\section{Experiments}
In this section, we first present details of training and testing the proposed method. Then, we present the results on the Sim2Real cross-domain generalizability, followed by showing results on the adversarial robustness. \textbf{Our PyTorch source code is provided in the supplementary material.}

\subsection{Settings and Implementation Details}
\textbf{Data.} We evaluate our method on the SceneFlow~\cite{dispnet16} and KITTI2015~\cite{Menze2015CVPR} datasets. The SceneFlow dataset is a large-scale synthetic dataset that contains $35,454$ training images and $4,370$ test images at the resolution of $540\times960$. Since it provides dense ground-truth disparities, it is widely used for pretraining DNN-based stereo matching methods. The KITTI2015 dataset is a real-world dataset of driving scenes, which contains 200 training images and 200 test images at the resolution of $375\times1242$. Since the depth of each scene is obtained through LiDAR, the ground truth is not dense. In addition, we also test pretrained models on the KITTI2012~\cite{Geiger2012CVPR} and the Middlebury~\cite{scharstein2014high} dataset at quarter resolution.

\textbf{Implementation Details.} Our method is implemented in PyTorch and trained end-to-end using the Adam optimizer with $\beta_1=0.9$ and $\beta_2=0.999$. All images are preprocessed with color normalization. During training, we use a batch size of 8 on four GPUs (Tesla V100) using $240\times576$ random crops from the input images. The maximum disparity level is set to $192$ and any values larger than this threshold will be ignored during training. 
For SceneFlow, we train our model from random initialization for 20 epochs with a constant learning rate of $0.001$. For KITTI2015, we split the $200$ training images into a training set of 140 images and a validation set of 60 images. We fine-tune our model pretrained on SceneFlow for another 600 epochs and use the validation set to select the best model. If no feature backbone is used to extract context information from the left image (Fig.~\ref{fig:model}), our model is denoted as \textbf{``ours w/o backbone"} or \textbf{``ours w/o b."} in tables and figures. 

To compare with adversarial training, we fine-tuned each method on \textit{KITTI2015} training images perturbed by 3-step unconstrained PGD attacks for 20 epochs, denoted as \textbf{adv.} in tables.

\textbf{Evaluation Metrics.} We adopt the provided protocols in the two datasets. There are three metrics: \textbf{EPE [px] } which measures the end-point error in pixels, \textbf{Bad 1.0 [\%]}  and \textbf{Bad 3.0 [\%]} which represents the error rate of errors larger than 1 pixel and 3 pixels respectively. 

\textbf{Baseline Methods.} We compare with state-of-the-art deep stereo matching methods: the PSMNet \cite{chang2018pyramid}, the GANet~\cite{GANet}, and the LEAStereo~\cite{LEAStereo}. We use their publicly released codes and trained model checkpoints in comparisons. 

\begin{table*}
\begin{center}
{\small
\begin{tabular*}{0.95\textwidth}{@{\extracolsep{\fill}}lccccccccccc}
\toprule
&\multicolumn{3}{c}{KITTI 2015} && \multicolumn{3}{c}{KITTI 2012} && \multicolumn{3}{c}{Middlebury}  \\ 
\cline{2-4} \cline{6-8} \cline{10-12}
Models (trained on SceneFlow) & EPE & Bad 1.0 & Bad 3.0 && EPE & Bad 1.0 & Bad 3.0 && EPE & Bad 1.0 & Bad 3.0\\
\midrule

PSMNet & 6.89 & 72.93 & 31.55 && 5.90 & 71.59 & 28.42 && 4.33 & 73.01 & 19.01\\
GANet & 1.66 & 42.12 & 10.48 && 1.48 & 31.61 & 9.51 && 2.26 & 27.45 & 11.40\\
LEAStereo & 2.00 & 51.29 & 13.90 && 1.91 & 44.26 & 14.28 && 3.47 & 32.67 & 14.81\\
\cline{2-4} \cline{6-8} \cline{10-12}
Ours w/o backbone & \textbf{1.25} & \textbf{25.95} & \textbf{6.12} && \textbf{1.23} & \textbf{19.66} & \textbf{6.80} && \textbf{1.71} & \textbf{18.72} & \textbf{9.16} \\
Ours & 1.26 & 27.92 & 6.31 && 1.28 & 20.62 & 7.16 && 1.96 & 20.09 & 10.05 \\
\bottomrule
\end{tabular*}}
\end{center}
\caption{Comparisons for the Sim2Real cross-domain generalizability from the SceneFlow trained models to the KITTI 2015, KITTI 2012 and Middleburry datasets in testing without any fine-tuning.}
\label{table:cross_domain}
\end{table*}

\begin{table*} [t]
\begin{center}
{\small
\begin{tabular*}{0.95\textwidth}{@{\extracolsep{\fill}}lllllllllllllll}
\toprule
& \multicolumn{4}{c}{EPE} & &\multicolumn{4}{c}{Bad 1.0} & &\multicolumn{4}{c}{Bad 3.0}   \\
\cline{2-5} \cline{7-10} \cline{12-15}
&CL &CT &CT  &UCT &  &CL &CT  &CT  &UCT  &&CL &CT  &CT  &UCT \\
& &$0.03$ &$0.06$ &$0.03$ & & &$0.03$ &$0.06$ &$0.03$ & & &$0.03$ &$0.06$ &$0.03$\\
\midrule
PSMNet & 0.28 & 29.05 & 84.04 & 91.08 &&2.00 & 84.75 & 90.41 & 92.75 && 0.16 & 54.80 & 83.68 & 89.91 \\
GANet & \textbf{0.25} & 3.93 & 9.75 & 23.75 && \textbf{1.42} & 70.64 & 84.68 &89.48 && \textbf{0.10} & 29.94 & 68.70 &79.11 \\
LEAStereo & 0.37 & 4.02 & 11.38 & 14.71 && 4.54 & 71.20 & 83.24 & 82.42 && 0.42 & 29.09 & 63.61 & 64.31 \\
\cline{2-5} \cline{7-10} \cline{12-15}

Ours w/o b. & 0.38 & 1.13 & 1.43 & 2.36 && 4.14 & 24.64 & 30.69 & 41.34 && 0.32 & \textbf{2.46} & 8.05 & 16.30 \\
Ours & 0.36 & \textbf{0.88} & \textbf{1.16} & \textbf{1.81} && 3.61 & \textbf{21.20} & \textbf{29.19} & \textbf{36.42} && 0.27 & 3.75 & \textbf{6.17} & \textbf{11.29} \\
\toprule
PSMNet + adv.  & 0.46 & 0.70 & 1.02 & 1.06 && 8.04 & 17.78 & 33.54 & 36.50 && 0.66 & 1.40 & 3.08 & 4.14 \\
GANet + adv. & 0.42 & 0.65 & 0.98 & 1.05 && 6.47 & 14.99 & 28.56 & 31.22 && 0.63 & 1.40 & 3.76 & 4.36 \\
LEAStereo + adv.  & 0.51 & 0.81 & 1.23 & 1.30 && 9.89 &21.73 &38.72 &42.06 && 0.99 &2.34 &5.59 &6.07 \\
\cline{2-5} \cline{7-10} \cline{12-15}
Ours w/o b. + adv.  & 0.42 &0.78 &0.90 &1.26 && 5.95 &16.83 &21.42 &32.27 && 0.73 &2.88 &3.83 &7.51 \\
Ours + adv.  & \textbf{0.41} &\textbf{0.61} &\textbf{0.69} &\textbf{0.88} && \textbf{5.77} &\textbf{13.46} &\textbf{16.29} &\textbf{22.93} && \textbf{0.52} &\textbf{1.39} &\textbf{2.00} &\textbf{3.99} \\
\bottomrule
\end{tabular*}}
\end{center}
\caption[PGD Attack Results in the KITTI2015 training dataset~\cite{Menze2015CVPR}.]{PGD Attack Results in the KITTI2015 training dataset~\cite{Menze2015CVPR}. For each metric, the four columns show that metric on \textit{CL}ean image, stereo-constrained attacked image (\textit{CT}, $\epsilon=0.03$), stereo-constrained attacked image (\textit{CT}, $\epsilon=0.06$), and unconstrained attacked image (\textit{UCT}, $\epsilon=0.03$). Note that on clean images, the results are performance on all the training and validation data, which are affected by different training-validation splits.}
\label{table:kitti15_attack_results}
\end{table*}

\begin{figure*} [t]
\begin{center}
\includegraphics[width=0.85\linewidth]{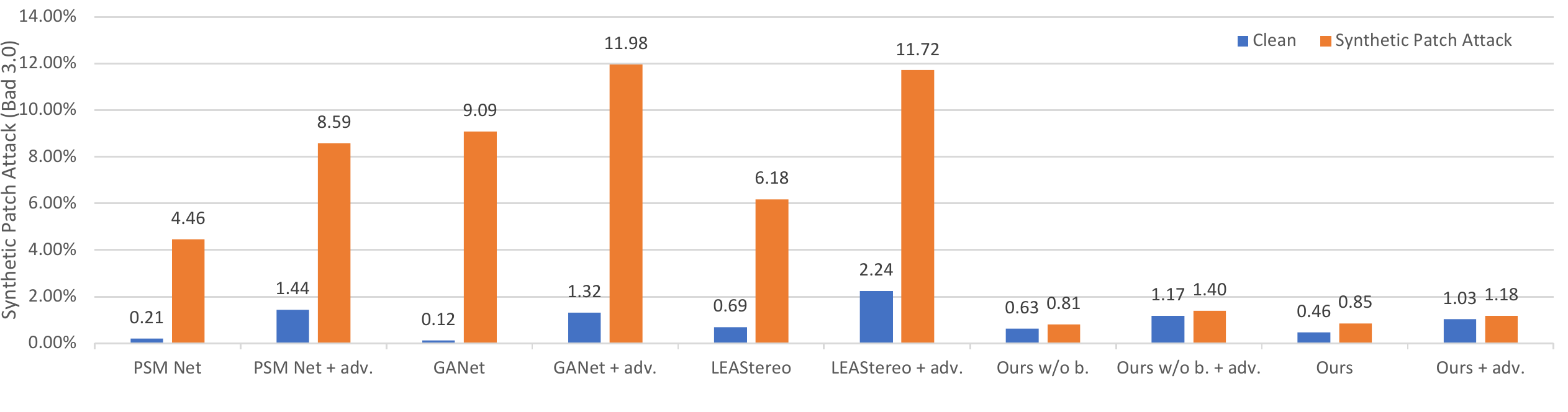}
\end{center}
   \caption{Adversarial Patch Attack Results in the KITTI2015 training dataset with photometric consistency retained in attack.}
\label{fig:patch_attack_kitti}
\end{figure*}

\begin{figure*} [t]
\begin{center}
\includegraphics[width=0.85\linewidth]{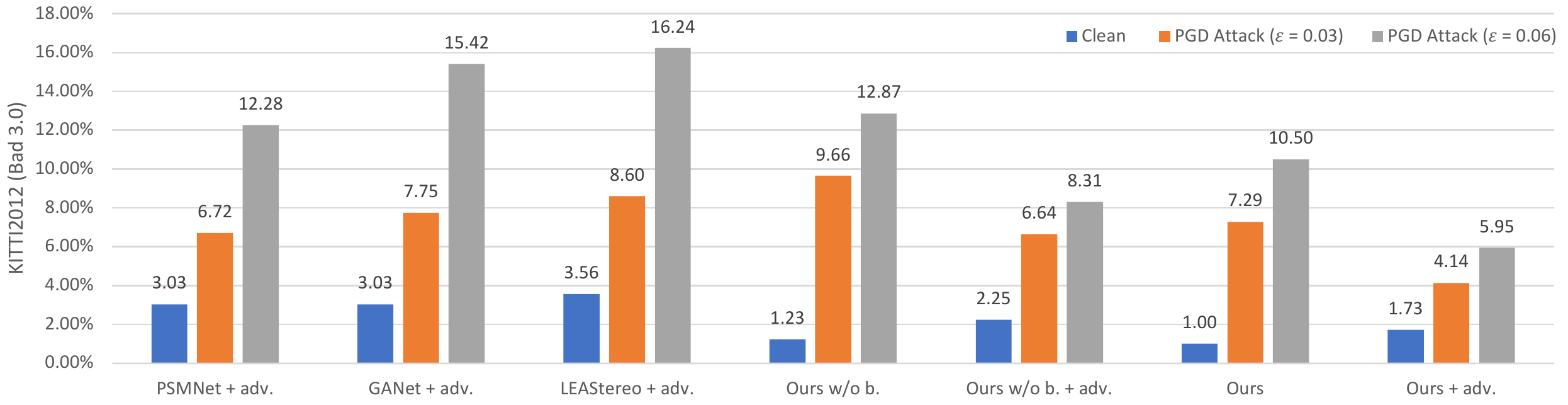}
\end{center}
   \caption{Transferrability of Adversarial Robustness: stereo-constrained 20-step PGD Attack Results in the KITTI2012 training dataset using adversarially trained neural networks on KITTI2015.}
\label{fig:trans_kitti12_attack_results}
\end{figure*}

\subsection{Sim2Real Cross-Domain Generalizability}
To verify the conjecture that cross-domain generalizability in stereo matching can be induced by removing the dependency between the cost volume computation and the dataset-dependent feature backbone, we evaluate all models pretrained on SceneFlow directly on the KITTI training datasets and the Middlebury training dataset~\cite{scharstein2014high}. As shown in \textbf{Table \ref{table:cross_domain}}, \textbf{our method outperforms prior art by a large margin}. This result shows that our proposed design of combining a non-parametric cost volume formed by the multi-scale census transform and a generalized cost aggregation/optimization DNN is indeed more cross-domain consistent. It also shows that the head sub-network DNN indeed learns to play the role of a domain-independent optimizer over a given cost volume.

\subsection{Results in KITTI}
\textbf{Experiment I): Adversarial Robustness Comparisons}. To evaluate the adversarial robustness in KITTI2015, we directly test the trained models on the entire training dataset (200 images).
Due to the GPU memory limitation, we only use the $240\times384$ center part of each image. Because of cropping, we also ignore those pixels where their correspondences are outside of the cropped images. We test both the stereo-constrained attack with $\epsilon=0.03,0.06$ and the unconstrained attack with $\epsilon=0.03$.  \textbf{Table~\ref{table:kitti15_attack_results}} shows the comparison results.

From the results, we show that state-of-the-art stereo matching methods are indeed vulnerable against adversarial attacks, even when photometric-consistency is preserved. Such vulnerability may raise serious concerns for the deployment of DNNs in safety critical applications. In contrast, \textbf{our method shows significantly better robustness on both stereo-constrained and unconstrained attacks.}

Note that our approach without feature backbone is unattackable if our proposed differentiable approximation of the comparison operator is not applied. Although this approach is highly non-linear, adversarial attacks can still find ways to perturb the input images, which further demonstrate the vulnerability of the DNNs. Interestingly, our method with the context feature backbone is more robust than its counterpart, showing that \textbf{the majority of the vulnerability actually comes from the matching part rather than the contextual information.} 
 
\textbf{Experiment II):  Adversarial Patch Attack.} To test if the adversarial vulnerability can be intentionally exploited in a more realistic setting, such as autonomous driving, we constructed the patch attack experiment to demonstrate the possibility of such attempts. We select $10$ scenarios where $40\times{40}$ adversarial patches can be put on more flat surfaces (\eg Fig. \ref{fig:teaser}). To preserve the depth of the scene, the ground truth disparities of the patches are the same as the corresponding part of the original image. For each image pair, we apply stereo-constrained PGD attacks with $100$ iterations. The \textbf{Bad 3.0} results are shown in \textbf{Fig.~\ref{fig:patch_attack_kitti}}. Note that the errors are computed using the whole image, while only a small portion of the image is affected. \textbf{Our method is very robust against adversarial patches. In contrast, other methods perform poorly, even with adversarial training.} This experiment also demonstrates the over-fitting tendency of adversarial training towards certain types of attack. More illustrations are shown in the supplementary materials.

\textbf{Experiment III): Comparisons with Adversarial Training.}
Our method increases the built-in adversarial robustness of stereo matching DNNs and thus it is orthogonal to existing defense methods such as adversarial training. From Table~\ref{table:kitti15_attack_results}, our method without adversarial training shows comparable adversarial robustness with $\epsilon=0.03,0.06$, especially for EPE and Bad $1.0$. For the patch attack experiment, our method is much more robust than others with adversarial training. \textbf{With adversarial training, our method has a stronger robustness than all other methods,} showing that our approach is indeed orthogonal to adversarial training and they can be jointly used to further improved robustness.

Besides adversarial robustness of the trained dataset KITTI2015, we also test on KITTI2012 to see how the adversarial robustness generalize on unseen data. In \textbf{Fig.~\ref{fig:trans_kitti12_attack_results}, our method shows a stronger cross-domain adversarial robustness than other adversarially trained methods.} Similarly, our method with adversarial training is still the most robust over all methods.

\textbf{Experiment IV): Ablation Study.} 
The census transform (CT) is chosen due to its non-differentiability and the fact that it is a well-rounded choice in the literature. We use multi-scale representations to respect the common recognition of its expressivity, and to alleviate choosing window size as a dataset-dependent hyper-parameter. We compare with traditional Sum of Absolute Difference (SAD) and show the results in \textbf{Table \ref{table:compare_descriptors}}. \textbf{We can see that CT is indeed much more robust than SAD due to its non-differentiability. CT with multiple scales has a stronger robustness than the single-scale version, while having a slightly better accuracy due to its flexibility. }

\begin{table} [ht]
\begin{center}
\resizebox{0.49\textwidth}{!}{
\begin{tabular}{|l c c| c c| c c|}
\hline
&\multicolumn{2}{c|}{SceneFlow} & \multicolumn{2}{c|}{KITTI15 (pretrained)} & \multicolumn{2}{c|}{KITTI15 Attack ($\epsilon=0.03$)}\\
Models & EPE [px] & Bad 3.0 [\%] & EPE [px] & Bad 3.0 & EPE [px] & Bad 3.0 [\%]\\
\hline
multi-scale SAD & 1.02 & 4.02 & 1.71 & 9.69 & 2.30 & 18.20\\
CT (w=11) & 1.18 & 4.77 & 1.28 & 6.38 & 1.88 & 7.22\\
\hline
Ours w/o backbone & 1.10 & 4.40 & \textbf{1.25} & \textbf{6.12} & 1.13 & \textbf{2.46} \\
Ours & \textbf{0.84} & \textbf{3.70} & 1.26 & 6.31 & \textbf{0.88} & 3.75 \\
\hline
\end{tabular}}
\end{center}
\caption{Comparison with CT with window size $11$ and multi-scale SAD}
\label{table:compare_descriptors}
\end{table}

\textbf{Experiment V): Leaderboard Comparisons.} \textbf{Table~\ref{table:kitti15_results_leaderboard}} shows the comparisons. Our method is slightly worse than state-of-the-art methods. As aforementioned, the detail of fine-tuning the pretrained model may play a significant role in the leaderboard comparisons. Our method is only fine-tuned by 140 training images in a vanilla manner without any bells and whistles. The gap may be bridged if more ablation studies are conducted to tune hyperparameters. 

\begin{table}[ht]
\begin{center}
\resizebox{0.4\textwidth}{!}{
\begin{tabular}{|l c c| c c|}
\hline
\quad Bad 3.0 [\%] &\multicolumn{2}{c|}{Non-Occlusion} & \multicolumn{2}{c|}{All Areas}\\
Models & FG & Avg All & FG & Avg All\\
\hline
GCNet~\cite{gcnet2017} & 5.58 & 2.61 & 6.16 & 2.87 \\
PSMNet~\cite{chang2018pyramid} & 4.31 & 2.14 & 4.62 & 2.32 \\
GANet-15~\cite{GANet} & 3.39 & 1.84 & 3.91 & 2.03 \\
GANet-Deep~\cite{GANet} & \textbf{1.34} & 1.63 & \textbf{1.48} & 1.81 \\
LEAStereo~\cite{LEAStereo} & 2.65 & \textbf{1.51} & 2.91 & \textbf{1.65} \\
\hline
Ours & 3.54 & 2.09 & 4.16 & 2.39 \\
\hline
\end{tabular}}
\end{center}
\caption{KITTI2015 leaderboard. FG: foreground regions.}
\label{table:kitti15_results_leaderboard}
\end{table}

\section{Conclusions}
This paper presents a novel workflow for stereo matching, which harnesses the best of classic features (multi-scale census transform) and end-to-end trainable DNNs for adversarially-robust and cross-domain generalizable stereo matching. The proposed method is motivated by the observation that DNN-based stereo matching methods can be deceived by a type of physically realizable attacks that entail stereo constraints in learning the perturbation. To address the adversarial vulnerability, this paper proposes to rethink the DNN feature backbone used in computing the cost volume by removing it for the matching stage. In experiments, the proposed method is tested in SceneFlow and KITTI2015 datasets with significantly better adversarial robustness and Sim2Real cross-domain generalizability (when no fine-tuning is used) achieved. It also obtains on-par performance on clean images.

{\small
\bibliographystyle{ieee_fullname}
\bibliography{main}
}

\end{document}